\documentclass[final]{cvpr}

\usepackage{times}
\usepackage{epsfig}
\usepackage{graphicx}
\usepackage{amsmath}
\usepackage{amssymb}

\usepackage{pifont}
\usepackage[pagebackref=true,breaklinks=true,colorlinks,bookmarks=false]{hyperref}

\def\cvprPaperID{****} 
\def\confYear{CVPR 2021}

\begin{document}

\title{MeNToS : Tracklets Association with a Space-Time Memory Network}

\author{Mehdi Miah, Guillaume-Alexandre Bilodeau and Nicolas Saunier\\
Polytechnique Montréal\\
{\tt\small \{mehdi.miah, gabilodeau, nicolas.saunier\}@polymtl.ca}}

\maketitle

\begin{abstract}

    We propose a method for multi-object tracking and segmentation (MOTS) that does not require fine-tuning or per benchmark hyperparameter selection. The proposed method addresses particularly the data association problem. Indeed, the recently introduced HOTA metric, that has a better alignment with the human visual assessment by evenly balancing detections and associations quality, has shown that improvements are still needed for data association. After creating tracklets using instance segmentation and optical flow, the proposed method relies on a space-time memory network (STM) developed for one-shot video object segmentation to improve the association of tracklets with temporal gaps. To the best of our knowledge, our method, named MeNToS, is the first to use the STM network to track object masks for MOTS. We took the $4^{th}$ place in the RobMOTS challenge. The project page is \url{https://mehdimiah.com/mentos.html}.

\end{abstract}

\section{Introduction}

Multi-object tracking (MOT) is a core problem in computer vision. Given a video, the objective is to detect all objects of interest then to track them throughout the video with consistent identities. Common difficulties are occlusions, small objects, fast motion (or equivalently low framerate) and deformations. 
Recently, the multi-object tracking and segmentation (MOTS) task~\cite{voigtlaender2019MOTSMultiObject} was introduced: instead of localizing objects with bounding boxes, they are described by their segmentation mask at the pixel level.

The MOTA metric has been commonly used to evaluate MOT but has a tendency to give more weight to detection errors compared to association errors. The newly introduced HOTA metric~\cite{luiten2020HOTAHigher} balances these two aspects and provides further incentives to work on the association step. That is why we developed a method which relies first on an instance segmentation followed by two data association steps. The first one is applied between consecutive frames using optical flow to obtain mask location prediction and mask Intersection over Union (mIoU) for matching. The second association step relies on a space-time memory network (STM). This is our main contribution. It is inspired by some results in one-shot video object segmentation (OSVOS), a task in computer vision that consists of tracking at a pixel level a mask provided only at the first frame. We use mask propagation with a STM network to associate tracklets separated by longer temporal gaps. Experiments show that the long-term data association significantly improves the HOTA score over the datasets used in the challenge.  

\begin{figure*}[ht]
    \centering               
    \includegraphics[scale = 0.48, trim={0 30 0 20},clip]{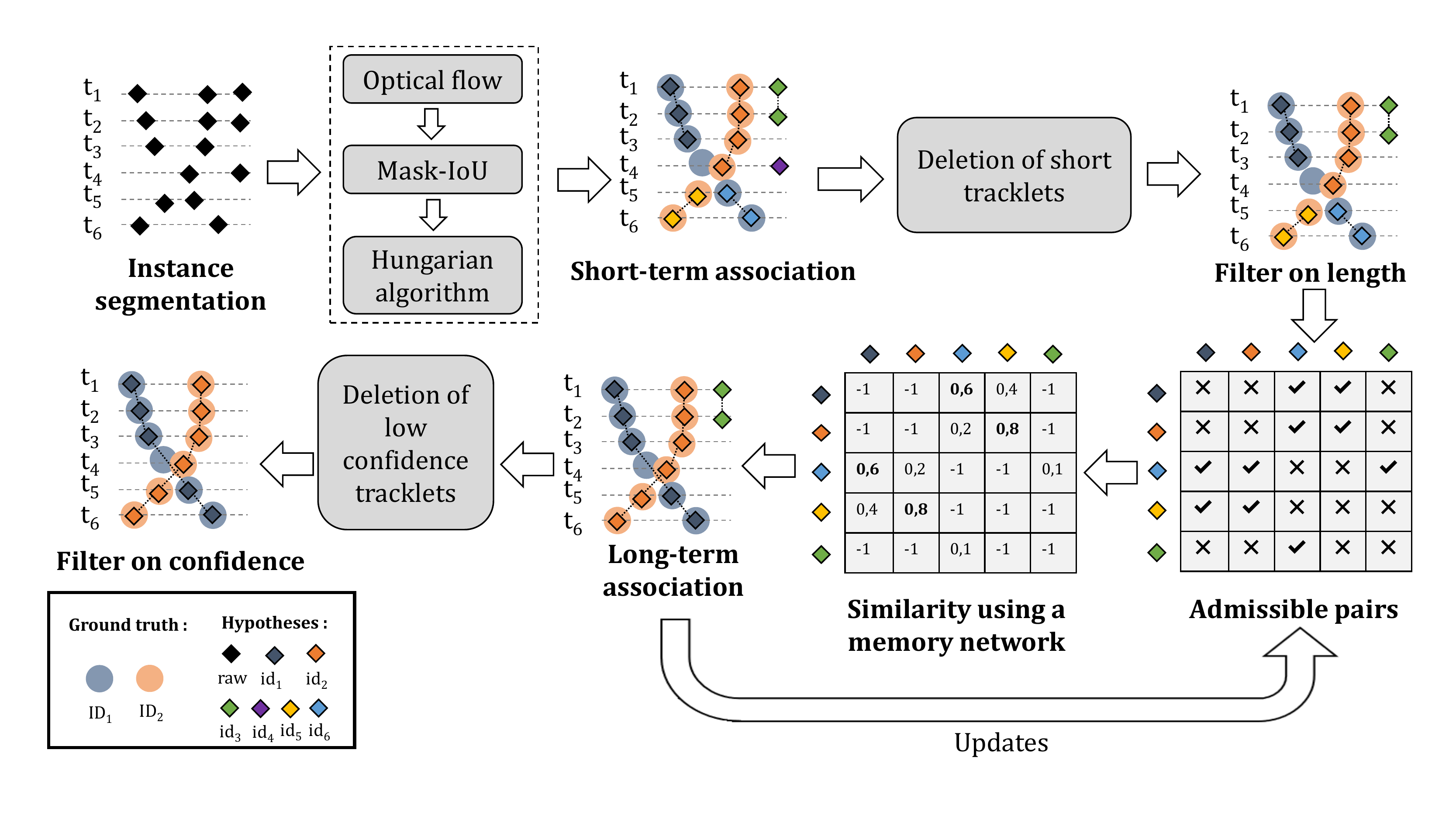}
    \caption{Illustration of our MeNToS method. Given an instance segmentation, binary masks are matched in consecutive frames to create tracklets. Very short tracklets are deleted. An appearance similarity, based on a memory network, is computed between two admissible tracklets. Then, tracklets are gradually merged starting with the pair having the highest similarity while respecting the updated constraints. Finally, low confidence tracks are deleted.}
    \label{fig:framework}
\end{figure*}


\section{Related works}

\paragraph{MOTS} Similarly to MOT where the ``tracking-by-detection'' paradigm is popular, MOTS is mainly solved by creating tracklets from segmentation masks then by building long-term tracks by merging the tracklets~\cite{luiten2020TrackReconstruct, yang2020ReMOTSSelfSupervised, zhang2020LIFTSLidar}.
Usually, methods use an instance segmentation method to generate binary masks; Re-MOTS~\cite{yang2020ReMOTSSelfSupervised} used two advanced instance segmentation methods and self-supervision to refine masks. As for the association step, many methods require a re-identification (reID) step. 
For instance, Voigtlaender et al.~\cite{voigtlaender2019MOTSMultiObject} extended the Mask R-CNN by an association head to return an embedding for each detection, Yang et al.~\cite{yang2020ReMOTSSelfSupervised} associated two tracklets if they were temporally close, without any temporal overlap with similar appearance features based on all their observations and a hierarchical clustering, while Zhang et al. ~\cite{zhang2020LIFTSLidar} used temporal attention to lower the weight of frames with occluded objects.

\paragraph{STM} Closely related to MOTS, OSVOS requires tracking objects whose segmentation masks are only provided at the first frame. STM~\cite{oh2019VideoObject} was proposed to solve OSVOS by storing some previous frames and masks in a memory that is later read by an attention mechanism to predict the new mask in a target image. Such a network was recently used~\cite{garg2021MaskSelection} to solve video instance segmentation, a problem in which no prior knowledge was given about the objects to track. However, it is unclear how STM behaves when multiple instances from the same class appear in the video. We show in this work they behave well and that they can help to solve a reID problem by taking advantage of the information at the pixel level and the presence of other objects.

\section{Method}

As illustrated in Figure~\ref{fig:framework}, our pipeline for tracking multiple objects is based on three main steps: detections of all objects of interest, a short-term association of segmentation masks in consecutive frames and a greedy long-term association of tracklets using a memory network.
    

\subsection{Detections}

Our method follow the ``tracking-by-detection'' paradigm. First, we used the public raw object masks provided by the challenge. They were obtained from a Mask R-CNN X-152 and Box2Seg Network. Objects with a detection score higher than $\theta_d$ and bigger than $\theta_a$ are extracted. Then, to avoid, for instance, that a car is simultaneously detected as a car and as a truck, segmentation masks having a mutual mIoU higher than $\theta_{mIoU}$ are merged to form a multi-class hypothesis. 

\subsection{Short-term association (STA)}

We associate temporally close segmentation masks between consecutive frames by computing the Farneback optical flow~\cite{farneback2003TwoFrameMotion} due to its simplicity. Masks from the previous frames are warped and a mIoU is computed between these warped masks and the masks from the next frame. 

The Hungarian algorithm is used to associate masks where the cost matrix is computed based on the negative mIoU. Matched detections with a mIoU above a threshold $\theta_s$ are connected to form a tracklet and the remaining detections form new tracklets. The class of the tracklet is the most dominant one among its detections.
Then, a non-overlap algorithm is applied to avoid any spatial overlap between masks, giving the priority to the pixels of the most confident mask. Finally, tracklets with only one detection are deleted since they often correspond to a false positive.

\subsection{Greedy long-term association (GLTA)}

GLTA and the use of a memory network for re-identification are the novelties of our approach. Once tracklets have been created, it is necessary to link them in case of fragmentation caused, for example, by occlusion. In this long-term association, we used a memory network to propagate some masks of a tracklet in the past and the future. In case of a spatial overlap with another tracklet, the two tracklets are merged. Given the fact that this procedure is applied at the pixel-level on the whole image, the similarity is only computed on a selection of admissible tracklets pairs to reduce the computational cost. At this step, we point out that all tracklets have a length longer than or equal to two. 

\subsubsection{Measure of similarity between tracklets}

\begin{figure}[ht] \centering
    \includegraphics[scale=0.4, trim={0 0 0 0},clip]{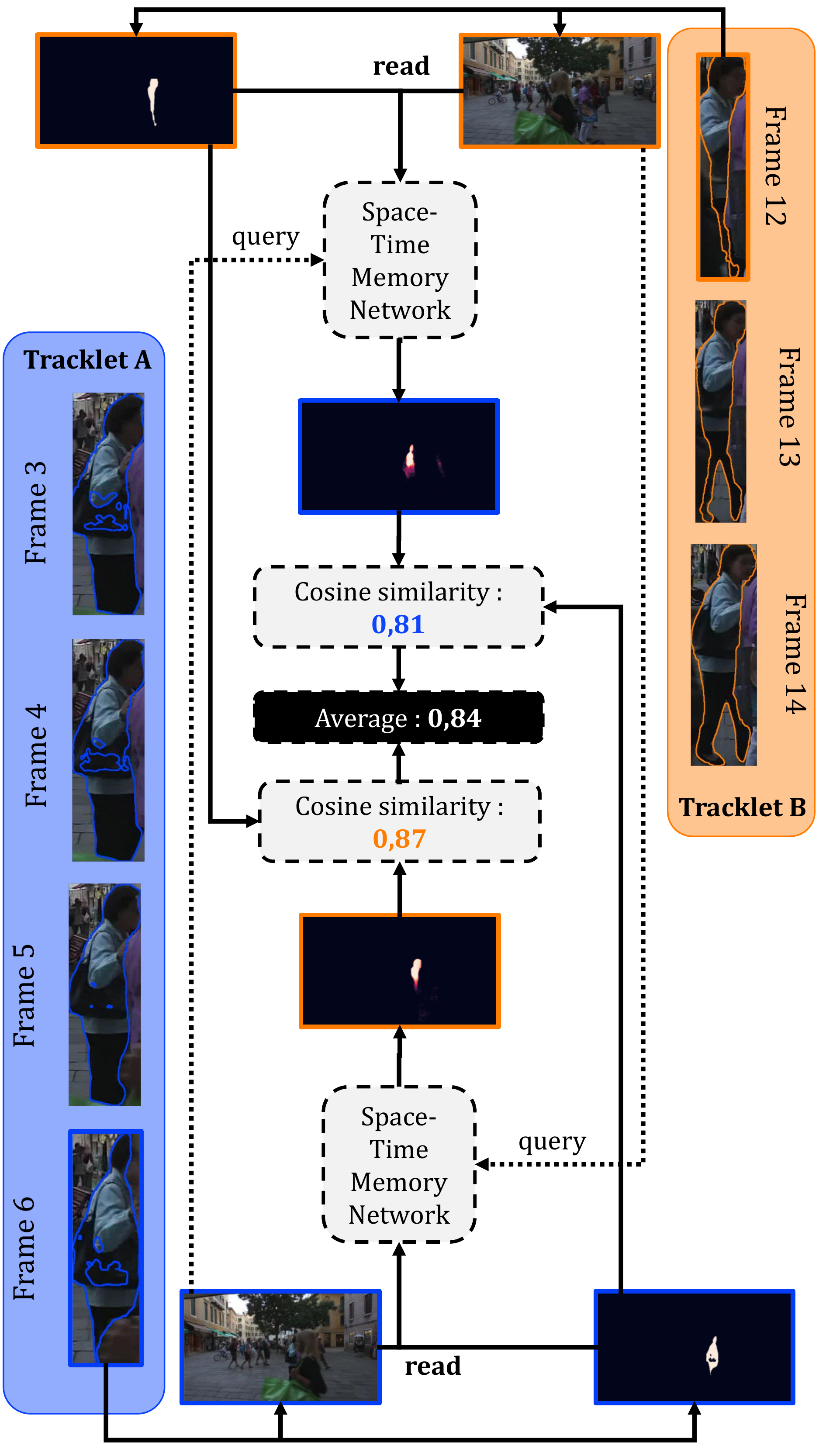}
    \caption{Similarity used at the long-term association step. For a matter of simplicity, only one mask and frame is used as reference and as target in the space-time memory network.}
    \label{fig:similarity}
\end{figure}

 Our similarity measure is based on the ability to match some parts of two different tracklets (say $T^A$ and $T^B$) and can be interpreted as a pixel-level visual-spatial alignment rather than a patch-level visual alignment~\cite{yang2020ReMOTSSelfSupervised, zhang2020LIFTSLidar}. For that, we propagate some masks of tracklet $T^A$ to other frames where the tracklet $T^B$ is present and then compare the masks of $T^B$ and the propagated version of the mask heatmaps, computed before the binarization, for $T^A$. The more they are spatially aligned, the higher the similarity is. In details, let us consider two tracklets $T^A = (M_1^{A}, M_2^{A}, \cdots, M_{N}^{A})$ and $T^B = (M_1^{B}, M_2^{B}, \cdots, M_P^{B})$ of length $N$ and $P$ respectively, such that $T^A$ appears first and where $M_1^A$ denotes the first segmentation mask of the tracklet $T^A$. We use a pre-trained STM network~\cite{oh2019VideoObject} to store two binary masks as references (and their corresponding frames): the closest ones ($M_N^A$ for $T^A$ and $M_1^B$ for $T^B$) and a second mask a little farther ($M_{N-n-1}^A$ for $T^A$ and $M_n^B$ for $T^B$). The farther masks are used because the first and last object masks of a tracklet are often incomplete due, for example, to occlusions. Then, the reference frames are used as queries to produce heatmaps with values between 0 and 1 ($H_N^A$, $H_{N-n-1}^A$, $H_1^B$, $H_n^B$). Finally, the average cosine similarity between these four heatmaps and the four masks ($M_N^A$, $M_{N-n-1}^A$, $M_1^B$, $M_n^B$) is the final similarity between the two tracklets. Figure~\ref{fig:similarity} illustrates a one-frame version of this similarity measure between tracklets. 

\subsubsection{Selection of pairs of tracklets} 

Instead of estimating a similarity measure between all pairs of tracklets, a naive selection is made to reduce the computational cost. The selection is based on the following heuristic: two tracklets may belong to the same objects if they belong to the same class, are temporally close, spatially close and with a small temporal overlap. 

In details, let us denote $f(M)$ the frame where the mask $M$ is present, $\bar{M}$ its center and $fps, H$ and $W$ respectively the number of frames per second, height and width of the video. The temporal ($C_t(T^A, T^B)$), spatial ($C_s(T^A, T^B)$) and temporal overlap ($C_o(T^A, T^B)$) costs between $T^A$ and $T^B$ are defined respectively as: 

\begin{equation}
    C_t(T^A, T^B) = \dfrac{|f(M_{N}^{A}) - f(M_1^{B})|}{fps},
\label{eq:Ct}
\end{equation}
\begin{equation}
    C_s(T^A, T^B) = \dfrac{2}{H+W} \times || \bar{M_N^A} - \bar{M_1^B} ||_1,
\label{eq:Cs}
\end{equation}
\begin{equation}
C_o(T^A, T^B) = |\{f(M)\, \forall M \in T^A\} \cap \{f(M)\, \forall M \in T^B\}|
\label{eq:Co}
\end{equation}

A pair $(T^A, T^B)$ is admissible if the tracklets belong to the same class, $C_t(T^A, T^B) \leq \tau_t$, $C_s(T^A, T^B) \leq \tau_s$ and $C_o(T^A, T^B) \leq \tau_o$.

\subsubsection{Greedy association}

 Similarly to Singh et al.~\cite{singh2017GreedyData}, we gradually merge the admissible pairs with the highest cosine similarity, if it is above a threshold $\theta_l$, while continuously updating the admissible pairs using equation \ref{eq:Co}. A tracklet can therefore be repeatedly merged with other tracklets.
Finally, tracks having their highest detection score lower than 90~\% are deleted.

\section{Experiments}
\subsection{Implementation details}

At the detection step, $\theta_d$ is 0.5, small masks whose area is less than $\theta_a=128$ pixels are removed and $\theta_{mIoU}=0.5$. 
For the GLTA step, the selection is done with $(\tau_t, \tau_s, \tau_o) = (1.5, 0.2, 1)$. To measure similarity, the second frame is picked using $n=5$. If that frame is not available, $n=2$, is used instead.
As for the thresholds at the STA and GLTA steps, we selected $\theta_s = 0.15$ and $\theta_l = 0.30$. These hyperparameters were selected through cross-validation and remain fixed regardless of the dataset and object classes. 

\begin{table*}[t]
\begin{center}
\begin{tabular}{|c|p{0.8cm}p{0.8cm}p{0.8cm}p{1.2cm}p{0.8cm}p{0.8cm}p{0.8cm}p{1.2cm}|p{0.7cm}p{0.7cm}p{0.7cm}|} 
\hline
Method & BDD & DAVIS & KITTI & MOTSCha. & OVIS & TAO & Waymo & YT-VIS & \multicolumn{3}{c|}{RobMOTS}\\
       & HOTA & HOTA & HOTA & HOTA & HOTA & HOTA & HOTA & HOTA & HOTA & DetA & AssA\\
\hline\hline
RobTrack~\cite{wei2021RobTrackRobust}      & \textcolor{red}{57.9} & \textcolor{red}{56.9} & \textcolor{blue}{71.6} & 61.0 & \textcolor{red}{61.6} & \textcolor{red}{55.0} & \textcolor{red}{57.2} & \textcolor{red}{68.3} & \textcolor{red}{61.2} & \textcolor{red}{59.4} & \textcolor{red}{64.8}\\
SBT~\cite{tang2021SBTSimple}           & 53.0 & \textcolor{blue}{50.3} & \textcolor{red}{74.0} & \textcolor{red}{64.4} & \textcolor{blue}{55.6} & \textcolor{blue}{51.8} & \textcolor{blue}{55.2} & \textcolor{blue}{64.4} & \textcolor{blue}{58.6} & \textcolor{blue}{55.9} & \textcolor{blue}{63.1}\\
SIA~\cite{ryu2021SIASimple}      & \textcolor{blue}{53.4} & 47.4 & 70.8 & \textcolor{blue}{62.2} & 54.8 & 49.6 & 54.1 & 62.7 & 56.9 & 55.8 & 59.8\\
MeNToS       & 52.3 & 49.6 & 69.7 & 60.2 & \textcolor{blue}{55.6} & 39.2 & 53.4 & 64.2 & 55.5 & 52.4 & 60.8\\
STP~\cite{luiten2021RobMOTSBenchmark}           & 49.4 & 48.2 & 66.4 & 60.4 & 52.8 & 43.8 & 51.8 & 62.3 & 54.4 & 55.8 & 55.0\\
\hline
\end{tabular}
\end{center}
\caption{Results on the RobMOTS test set. The HOTA metrics on each benchmark is reported alongside with the overall DetA, AssA and HOTA. \textcolor{red}{Red} and \textcolor{blue}{blue} indicate respectively the first and second best methods.}
\label{tab:results}
\end{table*}

\subsection{Datasets and performance evaluation}

The tracking algorithms are applied on the benchmarks on the RobMOTS challenge~\cite{luiten2021RobMOTSBenchmark}. It consists of eight tracking datasets with a high diversity in terms of framerate (ranges from 1 to 30 frames per second), objects of interest, duration and number of objects. Here, we considered the 80 categories of objects from COCO.



Recently, the HOTA metric was introduced to fairly balance the quality of detections and associations. It can be decomposed into the DetA and AssA metrics to measure the quality of these two components. The higher the HOTA is, the more the tracker is aligned with the human visual assessment. The final HOTA on RobMOTS is the average of the eight HOTA metrics.

\subsection{Results}


Results of Table~\ref{tab:results} indicate that our method is competitive for MOTS. MeNToS performs well on all benchmarks except TAO. This benchmark is more difficult for MeNToS, and all the other methods, since it is composed of videos with a very low framerate (1~fps). Without this outlier, MeNToS would perform on par with SIA. 

This issue comes from the second step of our method where the optical flow struggles to correctly associate consecutive masks of the same object. As a result, the deletion of very short tracklets leads to remove detections in this case, thus reducing DetA, the quality of detection (-3 percentage points on DetA compared to the baseline STP). 

However, MeNToS is able to correctly associate tracklets) in the long-term data association partially balances this drawback (+6 percentage points on AssA compared to STP).

\section{Conclusion}

In this work, we have developed a memory network-based tracker for multi-object tracking and segmentation. After creating tracklets, the STM network is used to compute a similarity score between tracklets. We can interpret this evaluation as a pixel-level visual-spatial alignment leveraging segmentation masks and the information of the whole image. Improving the creation of tracklets during the short-term data association may lead to further improvements.

\section*{Acknowledgment}
We acknowledge the support of the Natural Sciences and Engineering Research Council of Canada (NSERC), [DGDND-2020-04633 and DG individual 06115-2017]. 

{\scriptsize
\bibliographystyle{ieee_fullname}
\bibliography{mybiblio}
}

\end{document}